\documentclass[final]{cvpr}
\usepackage{url}
\usepackage{graphicx}
\usepackage{tabularx}
\usepackage{comment}
\usepackage{makecell}
\usepackage{pifont}
\usepackage{multirow}
\usepackage{subfigure}
\usepackage[misc]{ifsym}
\usepackage{afterpage}
\usepackage{setspace}
\usepackage{breakcites}
\usepackage{seqsplit}

\usepackage{times}
\usepackage{epsfig}
\usepackage{graphicx}
\usepackage{amsmath}
\usepackage{amssymb}

\usepackage{booktabs}

\usepackage[title]{appendix}

\usepackage{wrapfig}
\usepackage{hyphenat}
\usepackage{soul,color}
\usepackage{amsopn}
\usepackage{amsmath}
\usepackage{amssymb}
\usepackage{tabularx}
\usepackage{subfigure}

\definecolor{Red}{cmyk}{0,1,1,0}
\definecolor{Green}{cmyk}{1,0,1,0}
\definecolor{Cyan}{cmyk}{1,0,0,0}
\definecolor{Purple}{cmyk}{0.45,0.86,0,0}
\definecolor{Rosolic}{cmyk}{0.00,1.00,0.50,0}
\definecolor{Blue}{cmyk}{1.00,1.00,0.00,0}
\definecolor{BlueViolet}{cmyk}{0.86,0.91,0,0.04}
\definecolor{NavyBlue}{cmyk}{0.94,0.54,0,0}

\newcommand{\myparagraph}[1]{\vspace{0.1em}\noindent\textbf{#1}}

\usepackage[pagebackref=true,breaklinks=true,colorlinks,bookmarks=false]{hyperref}

\def\cvprPaperID{2693} 
\def\confYear{CVPR 2023}

\begin{document}

\title{End-to-End 3D Dense Captioning with Vote2Cap-DETR}

\author{
    Sijin Chen$^{1}$\thanks{This work is accomplished when visiting the Advanced Perception Reasoning Lab at I2R, A*STAR.} \quad\quad
    Hongyuan Zhu$^{2}$ \quad\quad
    Xin Chen$^{3}$ \quad\quad
    Yinjie Lei$^{4}$ \quad\quad
    Tao Chen$^{1}$\thanks{Corresponding author.} \quad\quad
    Gang YU$^{3}$
    \\
    $^{1}$Fudan University \quad
    $^{2}$Institute for Infocomm Research, A*STAR \\
    $^{3}$Tencent PCG \quad
    $^{4}$Sichuan University
    \\
    \tt \small \textbf{\href{https://github.com/ch3cook-fdu/Vote2Cap-DETR}{https://github.com/ch3cook-fdu/Vote2Cap-DETR}}
}



\maketitle

\begin{abstract}
%
3D dense captioning aims to generate multiple captions localized with their associated object regions. 
Existing methods follow a sophisticated ``detect-then-describe'' pipeline 
equipped with numerous hand-crafted components.
However, these hand-crafted components would yield suboptimal performance given cluttered object spatial and class distributions among different scenes. 
%
In this paper, we propose a simple-yet-effective transformer framework Vote2Cap-DETR based on recent popular \textbf{DE}tection \textbf{TR}ansformer (DETR). %
%
Compared with prior arts, our framework has several appealing advantages: 
1) Without resorting to numerous hand-crafted components,
our method is based on a full transformer encoder-decoder architecture with a learnable vote query driven object decoder, and a caption decoder that produces the dense captions in a set-prediction manner. 
2)  In contrast to the two-stage scheme, our method can perform detection 
and captioning in one-stage. 
3) Without bells and whistles, 
extensive experiments on two commonly used datasets, ScanRefer and Nr3D, demonstrate that our 
Vote2Cap-DETR surpasses current state-of-the-arts by 11.13\%
and 7.11\% in CIDEr@0.5IoU, respectively.
%
Codes will be released soon.
%

\end{abstract}
\section{Introduction}

\begin{figure}[htbp]
	\centering
	\includegraphics[width=\linewidth]{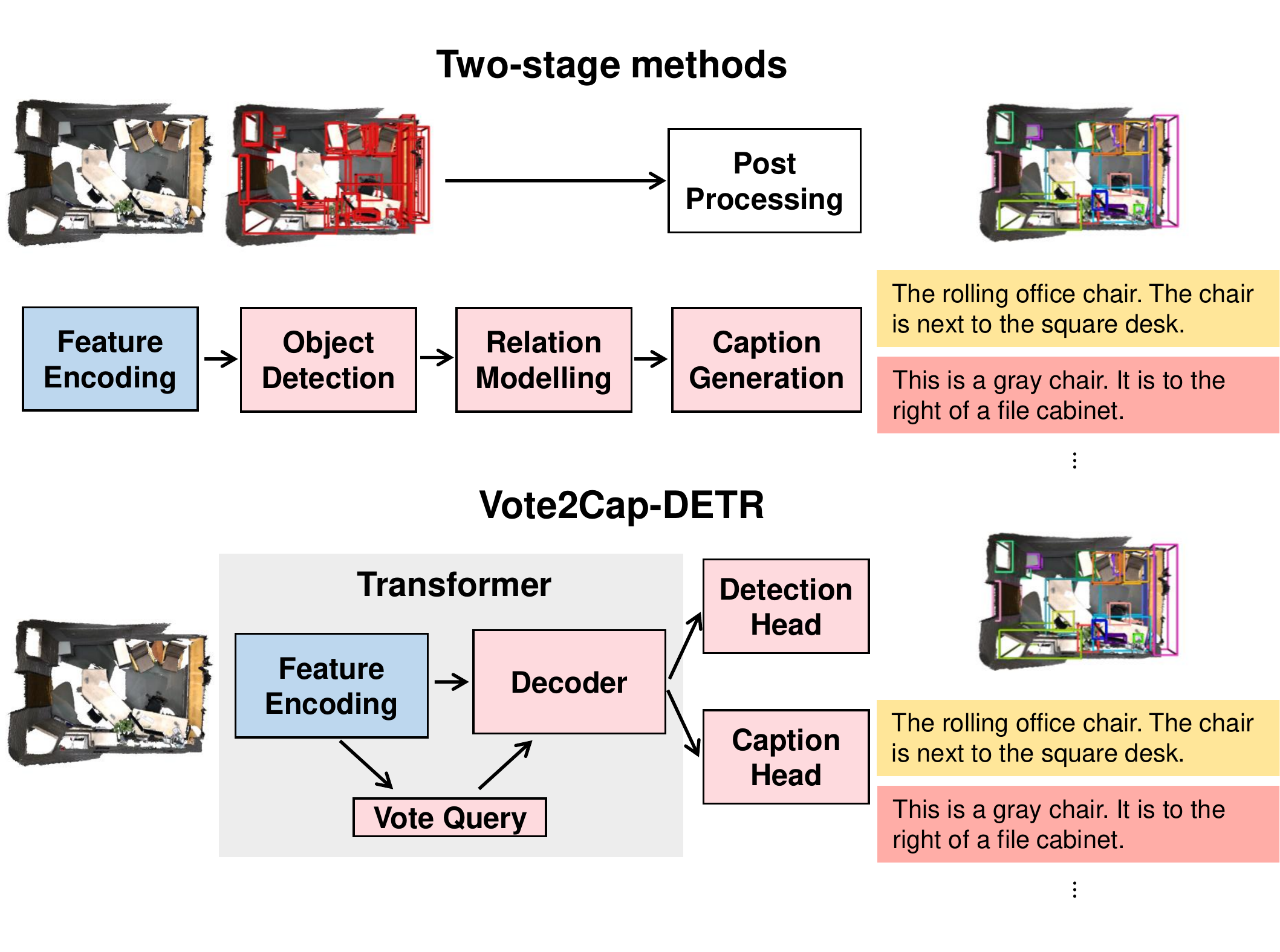}
	\caption{
	\textbf{Illustration of existing two-stage 3D dense captioning method (upper) and our Vote2Cap-DETR (bottom).}
	Existing methods adopt a two-stage pipeline that heavily depends on a detector's output.
	Therefore, we propose a transformer-based one-stage model, Vote2Cap-DETR, that frames 3D dense captioning as a set prediction problem.
	}\vspace{-10pt}
	\label{fig:teaser}
\end{figure}

3D dense captioning \cite{chen2021scan2cap,chen2021d3net,yuan2022x-trans2cap,wang2022spacap3d,jiao2022more,cai20223djcg} requires a system to localize all the objects in a 3D scene, and generate descriptive sentences for each object. 
%
%
This problem is challenging given 
1) the sparsity of point clouds and
%
2) the cluttered distribution of objects.

3D dense captioning can be divided into two tasks, object detection and object caption generation.
%
%
%
Scan2Cap\cite{chen2021scan2cap}, MORE\cite{jiao2022more}, and SpaCap3D\cite{wang2022spacap3d} propose well-designed relation reasoning modules to efficiently model relations among object proposals.
\cite{zhong2022contextual3DdenseCap} introduces contextual information from two branches to improve the caption.
%
%
3DJCG\cite{cai20223djcg} and D3Net\cite{chen2021d3net} study the correlation between 3D visual grounding and 3D dense captioning, and point out that these two tasks promote each other.
%
%
Additionally, $\chi$-Trans2Cap\cite{yuan2022x-trans2cap} discusses how to transfer knowledge from additional 2d information to boost 3d dense captioning.

Among existing methods, they all adopt a two-stage “detect-then-describe” pipeline\cite{chen2021scan2cap,jiao2022more,wang2022spacap3d,cai20223djcg,chen2021d3net,zhong2022contextual3DdenseCap} (Figure \ref{fig:teaser}).
This pipeline first generates a set of object proposals, then decodes each object by a caption generator with an explicit reasoning procedure.
Though these methods have achieved remarkable performance, the ``detect-then-describe'' pipeline suffers from the following issues: 
1) Because of the serial and explicit reasoning, this task highly depends on the object detection performance, which limits the mutual promotion of detection and captioning. 
2) The heavy reliance on hand-crafted components, e.g., radii, 3D operators, the definition of proposal neighbors, and post-processing (non-maximum suppression\cite{neubeck2006nms}) introduces additional hyper-parameters, leading to a sub-optimal performance given the sparse object surfaces and cluttered object distributions among different indoor scenes.
This inspires us to design an one-stage 3D dense captioning system.

To address the above issues, we propose Vote2Cap-DETR, a full transformer encoder-decoder architecture for one-stage 3D dense captioning.
Unlike the traditional ``detect-then-describe'' pipeline, we directly feed the decoder's output into the localization head and caption head in parallel.
By casting 3D dense captioning as a set-to-set problem, each target instance and its language annotation is matched with a query in an one-to-one correspondence manner, helping feature representation for proposals be more discriminative to identify each \textbf{distinctive} object in a 3D scene.
Additionally, we also propose a novel vote query driven decoder to introduce spatial bias for better localization of objects in a cluttered 3D scene.

With the fully attentional design, we resolve 3D dense captioning with the following innovations: 
1) Our method treats the 3D dense captioning task as a set prediction problem. The proposed Vote2Cap-DETR directly decodes the features into object sets with their locations and corresponding captions by applying two parallel prediction heads. 
2) We propose a novel vote decoder by reformulating the object queries in 3DETR into the format of the vote query, which is a composition of the embeddings of the seeds point and the vote transformation of the box with respect to the seeds. This indicates the connection between the vote query in Vote2Cap-DETR with the VoteNet, but with better localization and higher training efficiencies; 
3) We develop a novel query driven caption head, which absorbs the relation and attribute modeling into the self- and cross-attention, so that it can look into both the local and global context to better describe the scene.
Extensive experiments on two commonly used datasets, ScanRefer and Nr3D, demonstrate that our approach surpasses prior arts with many hand-crafted procedures by a large margin, which demonstrates the superiority that, full transformer architecture with sophisticated vote head and caption head can inspire many 3D vision and language tasks. 

To summarize, the main contributions of this work include:
\begin{itemize} 
\setlength\itemsep{0em}
    \item We propose a novel one-stage and fully attention driven architecture for 3D dense captioning as a set-to-set prediction problem, which achieves object localization and caption generation in parallel. 

	\item Extensive experiments show that our proposed Vote2Cap approach achieves a new state-of-the-art performance on both Nr3D\cite{achlioptas2020referit3d} (45.53\% C@0.5) and ScanRefer\cite{chen2021scan2cap} (73.77\% C@0.5).
	
\end{itemize} 
\section{Related Work}
We briefly summarize works on 3D dense captioning, and DETR-based methods for image and 3D object detection.
Additionally, we also introduce some methods for image captioning, which are closely related to our work.

\myparagraph{3D Dense Captioning.}
3D dense captioning, a task that requires translating 3D scene information to a set of bounding boxes and natural language descriptions, is challenging and has raised great interest among scholars recent years.
%
%
%
Scan2Cap\cite{chen2021scan2cap} and MORE\cite{jiao2022more} build graph on a detector's\cite{qi2019votenet,jiang2020pointgroup} box estimations with hand-crafted rules to reason complex relations among objects in a 3D scene.
%
SpaCap3D\cite{wang2022spacap3d} build a spatiality-guided transformer to model spatial relations among the detector's output.
3DJCG\cite{cai20223djcg} and D3Net\cite{chen2021d3net} study the joint promotion of 3D dense captioning and 3D visual grounding.
$\chi$-Trans2Cap\cite{yuan2022x-trans2cap} introduces additional 2D prior to complement information for 3D dense captioning with knowledge transfer.
Recently, \cite{zhong2022contextual3DdenseCap} shifts attention to contextual information for the perception of non-object information.
These approaches have made great attempts to solve the 3D dense captioning problem.
However, they all follow a ``detect-then-describe'' pipeline, which is heavily dependent on a detector's performance.
Our proposed Vote2Cap-DETR differs from existing works in that, our method is a one-stage model that detects and generates captions in parallel, and treats 3D dense captioning as a set prediction problem.

\myparagraph{DETR: from 2D to 3D.}
\textbf{DE}tection \textbf{Tr}ansformer(DETR)\cite{carion2020detr} is a transformer\cite{vaswani2017attention} based architecture that treats object detection as a set prediction problem, and does not require non-maximum suppression\cite{neubeck2006nms} for post-processing.
Though great results have been achieved, DETR suffers from slow convergence.
Many follow-up works\cite{zhu2020deformabledetr,zhang2022detr++,gao2021fast,meng2021conditionaldetr,chen2022groupdetr,jia2022hybriddetrs} put efforts on speeding up DETR's training by introducing multi-scale features, cross attention designs, and label assignment techniques.
Researchers also attempt to introduce transformer architectures to 3D object detection.
GroupFree3D\cite{liu2021groupfree3d} learns proposal features from the whole point cloud through the transformer rather than grouping local points.
3DETR\cite{misra2021-3detr} analyzes the potential of the standard transformer model, and generates proposals by uniformly sampling seed points from a 3D scene.
In our work, we extend the DETR architecture for 3D dense captioning that makes caption generation and box localization fully interrelated with parallel decoding.
Additionally, we propose vote query for better performance and faster convergence.

\myparagraph{Image Captioning.}
Image captioning requires a model to generate sentences describing key elements in an image, which has become a hot topic in computer vision.
Existing image captioning works adopt an encoder-decoder architecture, where the decoder generates sentences from visual features extracted by the encoder.
\cite{anderson2018bottom-up-top-down,cornia2020m2transformer,huang2019AinA-img-cap,pan2020x-linear-img-cap} adopt a detector to extract region features as visual clues for the decoder, while \cite{liu2021cptr,zhang2021rstnet} extract grid features directly from an image.
Additionally, \cite{nguyen2022grit} generates captions with both region and grid visual features.
%
%
%
Though these methods are effective in image captioning, they cannot be directly applied to 3D dense captioning, which requires both accurately localizing and describing a 3D object, rather than simply captioning a whole 2D scene image.
In contrast, our proposed caption head sufficiently leverages the rich context information in 3D point cloud, receives visual clues from both the object query and its local context, and fuses them to achieve effective 3D dense captioning.
\section{Method}
\begin{figure*}[htbp]
	\centering
	\includegraphics[width=\linewidth]{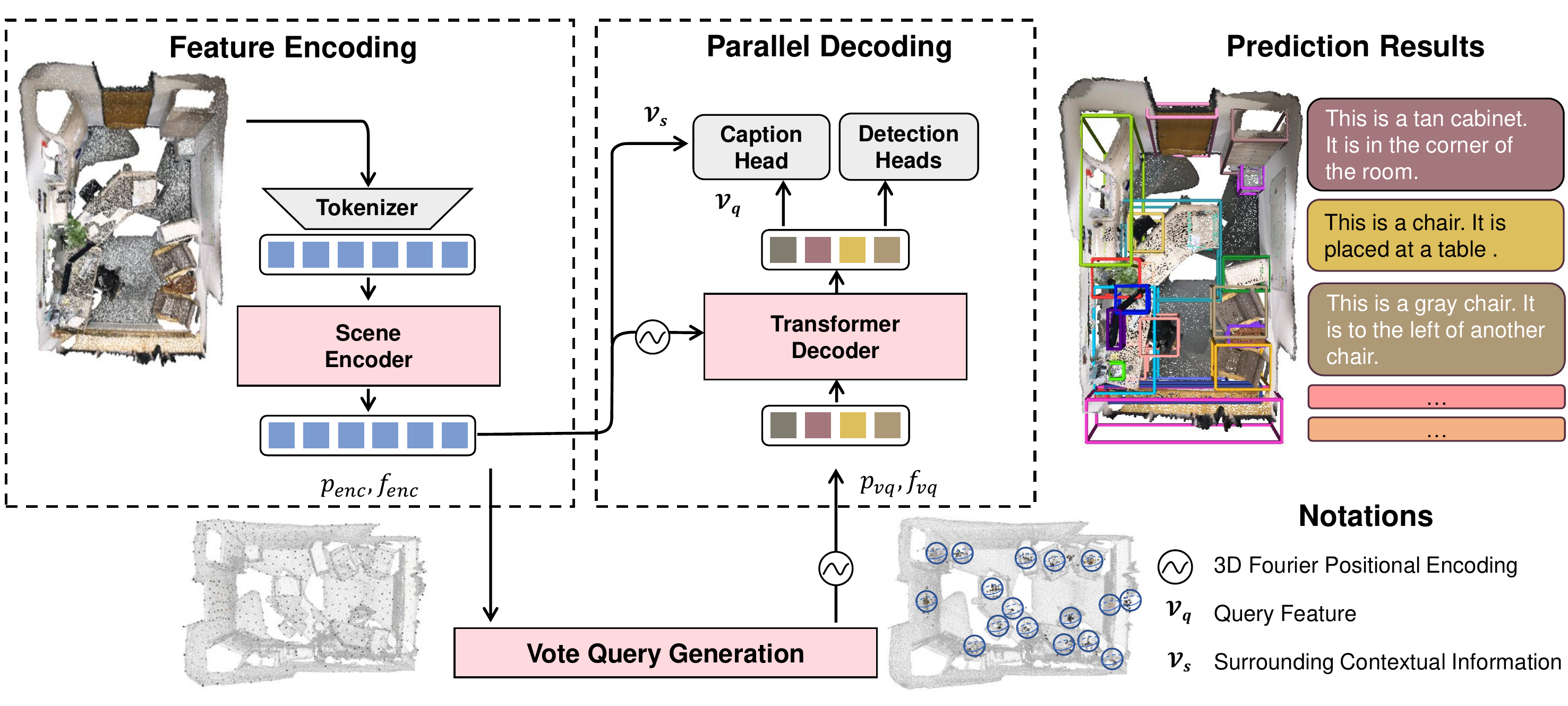}
	\caption{
	\textbf{Approach.} Vote2Cap-DETR is an one-stage transformer model that takes a 3D point cloud as its input, and generates a set of box predictions and sentences localizing and describing each object in the point cloud.
	The scene encoder first generates encoded scene tokens $(p_{enc}, f_{enc})$ from the input point cloud.
	Then, we generate vote query $(p_{vq}, f_{vq})$ from the encoded scene tokens, which introduce both spatial bias $p_{vq}$ and content-aware feature $f_{vq}$ to initial object queries.
	The transformer decoder decodes each vote query with two parallel task heads for captioning and detection.
	We optimize Vote2Cap-DETR with a set loss.
	}
	\label{fig:pipeline}
\end{figure*}

As shown in Fig. \ref{fig:pipeline}, given a 3D scene, our goal is to localize objects of interest and generate informative natural language descriptions for each object.
The \textbf{input} of our model is a point cloud $PC = \left[p_{in}; f_{in}\right] \in \mathbb{R}^{N \times \left(3 + F\right)}$ representing an indoor 3D scene.
Here, $p_{in} \in \mathbb{R}^{N\times 3}$ is the absolute locations for each point, and $f_{in} \in \mathbb{R}^{N\times F}$ is additional input feature for each point, such as \textit{color}, \textit{normal}, \textit{height}, or \textit{multiview feature} introduced by \cite{chen2021scan2cap, chen2020scanrefer}.
The expected \textbf{output} is a set of box-caption pairs $( \hat{B}, \hat{C} ) = \{(\hat{b}_1, \hat{c}_1), \cdots, (\hat{b}_K, \hat{c}_K)\}$, representing an estimation of $K$ distinctive objects in this 3D scene.

Specifically, our system adopts 3DETR\cite{misra2021-3detr} encoder as our scene encoder, and transformer decoder to capture both object-object and object-scene interactions by the attention mechanism. 
Then, we adopt two task-specific heads for object detection and caption generation. 

\subsection{3DETR Encoder}
Inspired by DETR\cite{carion2020detr}, 3DETR\cite{misra2021-3detr} has made a successful attempt at bringing full transformer architecture to the 3D object detection task, which removes many hard-coded design decisions as the popular VoteNet and PointNet++ modules in most two-stage methods.

In 3DETR encoder, the input $PC$ is first tokenized with a set-abstraction layer\cite{qi2017pointnet++}.
Then, point tokens are fed into a masked transformer encoder with a set-abstraction layer followed by another two encoder layers.
We denote the encoded scene tokens as $\left[p_{enc}; f_{enc}\right] \in \mathbb{R}^{1,024 \times \left(3+256\right)}$.
%
%
%
%
%

%
%
\subsection{Vote Query}
\begin{figure}[htbp]
	\centering
	\includegraphics[width=\linewidth]{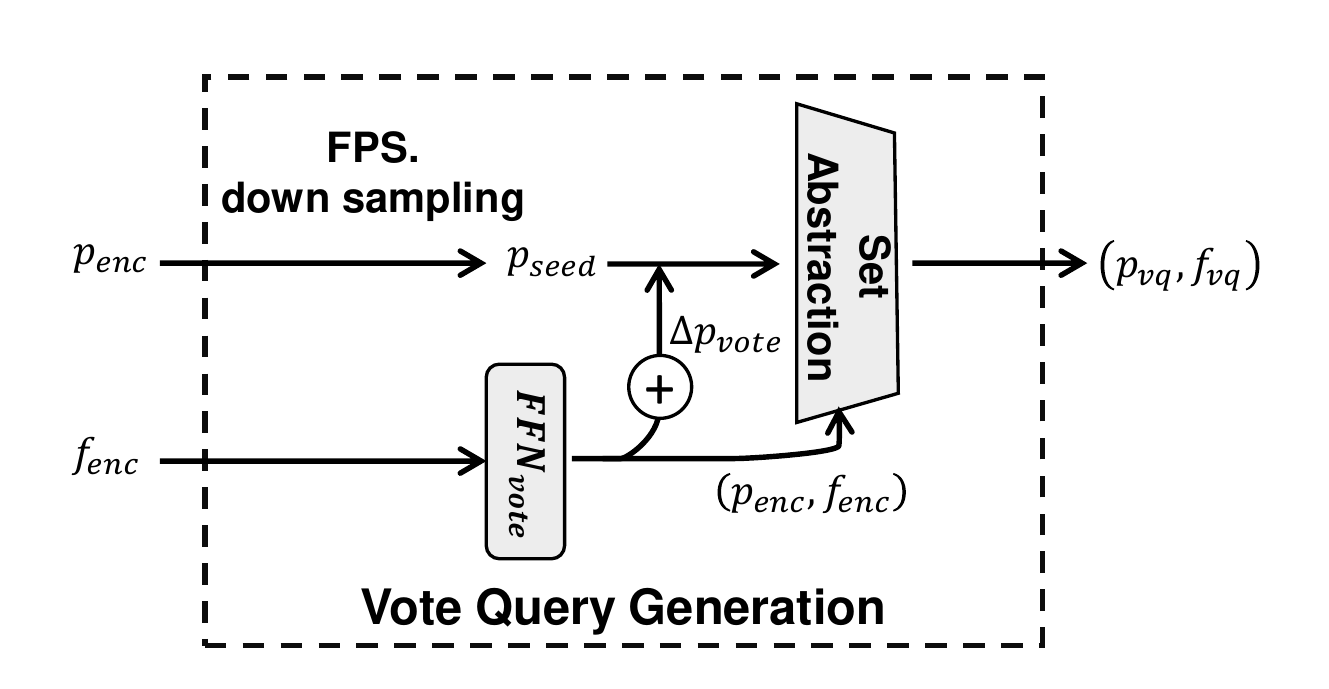}
	\caption{
        \textbf{Vote Query Generation.} 
        Vote query $p_{vq}$ contains spatial bias ($\Delta p_{vote}$) to initial object queries ($p_{seed}$), which are sampled from the scene with farthest point sampling (FPS) and gathered feature $f_{vq}$ from the point cloud for each query.
	}\vspace{-10pt}
	\label{fig:vote-query-generation}
\end{figure}

Though 3DETR has achieved initial success in 3D object detection, 
it suffers from certain limitations.
%
 3DETR proposes the box estimation around the query points (aka proposal centers) sampled from the scenes, which can make these boxes far away from real objects given the sparse object surfaces, resulting in slow convergence to capture discriminative object features with further miss detections.
%
%
%

Prior works on fast convergence DETR models\cite{meng2021conditionaldetr,chen2022conditionaldetrv2,zhang2022dino} show that by injecting more structured bias to initialize object queries, such as anchor points or content-aware queries, accelerates training.
%
%
Therefore, we propose the vote query, which introduces both 3D spatial bias and content-related information, for faster convergence and performance improvement. 
%

More specifically, we reformulate the object queries in 3DETR into the format of vote query, as a composition of the embedding of the reference points and vote transformation around them. This helps to build the connection between the object query in 3DETR and the vote set prediction widely studied in VoteNet. 
%
%

The detailed structure is shown in Figure \ref{fig:vote-query-generation}.
%
%
Here, vote $\Delta p_{vote}$ is predicted from encoded scene token feature $f_{enc}$ with a \textbf{F}eed \textbf{F}orward \textbf{N}etwork (FFN) $FFN_{vote}$ that learns to shift the encoded points to objects' centers spatially:
%
\begin{equation}
    p_{vote} = p_{enc} + \Delta p_{vote} = p_{enc} + FFN_{vote}\left(f_{enc}\right).
\end{equation}
Then, we sample 256 points $p_{seed}$ from $p_{enc}$ with farthest point sampling, and locate each point's offset estimation for $p_{vq} = p_{seed} + \Delta p_{vote}$.
Finally, we gather features from $(p_{enc}, f_{enc})$ for $p_{vq}$ with a set-abstraction layer\cite{qi2017pointnet++}, to formulate the vote query feature $f_{vq} \in \mathbb{R}^{256\times 256}$.
We represent vote query as $(p_{vq}, f_{vq})$.

Following 3DETR\cite{misra2021-3detr}, our model adopts an eight-layer transformer decoder, and the $i$-th layer's input query feature $f^{i}_{query}$ is calculated through
\begin{equation}
f^{i}_{query} = Layer_{i-1}\left(f^{i-1}_{query} + FFN\left(PE\left(p_{vq}\right)\right)\right),
\end{equation}
where $f^{0}_{query} = f_{vq}$, and $PE(\cdot)$ is the 3D Fourier positional encoding function\cite{tancik2020fourier}.
Experiments in later sections demonstrate that:
1) Vote query injects additional spatial bias to object detection and boosts the detection performance.
2) Encoding features from the point cloud as initial queries accelerates convergence.

\subsection{Parallel Decoding}
\label{sec:parrallel decoding heads}
We adopt two task-specific heads for simultaneous object detection and caption generation.
The two task heads are agnostic to each other's output.

\myparagraph{Detection Head.}
Detecting objects in a 3D scene requires box corner estimation $\hat{B}$ and class estimation $\hat{S}$ (containing ``no object'' class) from each object query feature.
Following 3DETR\cite{misra2021-3detr}, box corner estimation is reformulated into offset estimation from a query point to an object's center, and box size estimation.
All subtasks are implemented by FFNs.
In practice, the object localization head is shared through different layers in the decoder, following all existing works on DETR\cite{carion2020detr,misra2021-3detr,meng2021conditionaldetr,chen2022conditionaldetrv2}.

\myparagraph{Caption Head.}
\begin{figure}[htbp]
	\centering
	\includegraphics[width=\linewidth]{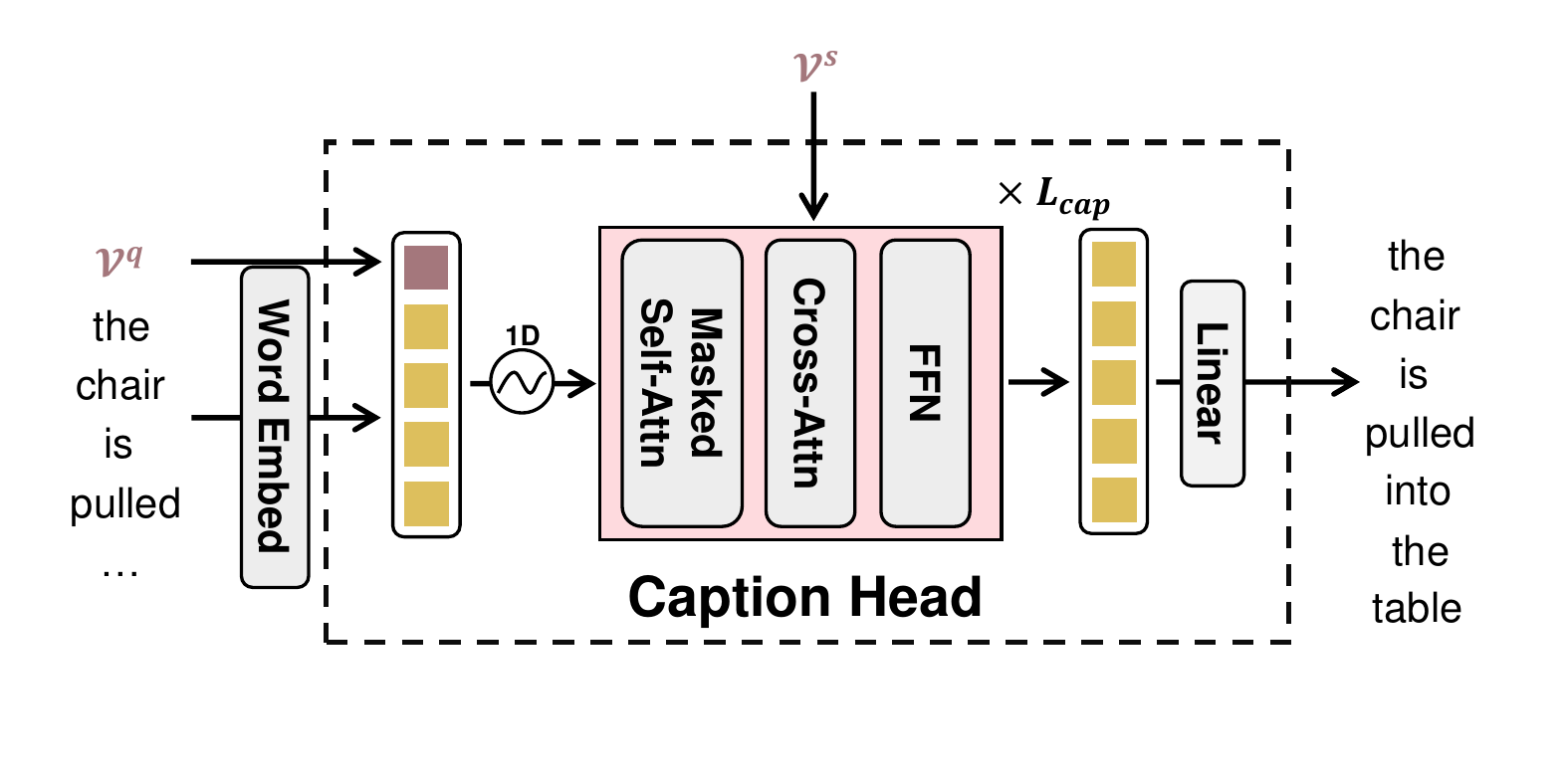}
	\caption{
	    \textbf{Dual-Clued Captioner(DCC).}
	    DCC is a lightweight transformer based caption head that uses vote query feature $\mathcal{V}_{q}$ as caption perfix to identify the described region, and contextual features $\mathcal{V}_{s}$ surrounding the vote query to complement with more surrounding information for more descriptive caption generation.
	}
	\label{fig:caption-head}\vspace{-10pt}
\end{figure}
%
3D dense captioning requires attribute details on an object and its relation with its close surroundings.
%
%
However, the vote query itself is agnostic to box predictions for the whole scene, and fails to provide adequate attribute and spatial relations for generating informative captions.
Therefore, the main difficulty is how to leverage sufficient surrounding contextual information without confusing the caption head.

To address the above issues, we propose \textbf{D}ual-\textbf{C}lued \textbf{C}aptioner(DCC), a lightweight transformer decoder-based caption head, for 3D dense captioning.
DCC consists of a stack of 2 identical transformer decoder blocks, sinusoid position embedding, and a linear classification head.
To generate informative captions, DCC receives two streams of visual clue $\mathcal{V}=(\mathcal{V}^{q}, \mathcal{V}^{s})$.
Here, $\mathcal{V}^{q}$ is the last decoder layer's output feature of a vote query, and $\mathcal{V}^{s}$ is contextual information surrounding the absolute location of each vote query.
When generating a caption for a proposal, we substitute the standard \textbf{S}tart \textbf{O}f \textbf{S}eqenece(`SOS') prefix with $\mathcal{V}^{q}$ of the described query identifying the object to be described following \cite{wang2022spacap3d}.
Since the vote query is agnostic of actual neighbor object proposals because of the parallel detection branch, we introduce the vote query's $k_{s}$ nearest local context token features as its local surroundings $\mathcal{V}^{s}$ as keys for cross attention.
During the evaluation, we generate captions through beam search with a beam size of 5.

\subsection{Set prediction loss for 3D Dense Captioning}
\label{sec:set-to-set-training}
Our proposed Vote2Cap-DETR generates a set of paired box-caption proposals $(\hat{B}, \hat{C})$ for 3D dense captioning.
It requires supervision for vote query ($\mathcal{L}_{vq}$), detection head ($\mathcal{L}_{det}$), and caption head ($\mathcal{L}_{cap}$).

\myparagraph{Vote Query Loss.}
We borrow vote loss from VoteNet\cite{qi2019votenet} as $\mathcal{L}_{vq}$, to help the vote query generation module learn to shift points $p_{enc}$ to an object's center:
\begin{equation}
    \mathcal{L}_{vq} = \frac{1}{M}
        \sum_{i = 1}^{M} \sum_{j = 1}^{N_{gt}}
            \left\|p_{vote}^{i} - cnt_{j}\right\|_{1}
                \cdot 
                \mathbb{I}\left\{p_{enc}^{i} \in I_j\right\}.
\end{equation}
Here, $\mathbb{I}(\cdot)$ is an indicator function that equals $1$ when the condition meets and $0$ otherwise, $N_{gt}$ is the number of instances in a 3D scene, $M$ is the size of $p_{vote}$, and $cnt_{j}$ is the center of $j$th instance $I_j$.

\myparagraph{Detection Loss.}
Following 3DETR\cite{misra2021-3detr}, we use the same Hungarian algorithm to assign each proposal with a ground truth label.
Since 3D dense captioning is closely related to the object localization ability, we apply a larger weight on the gIoU loss component for total set loss\cite{misra2021-3detr}:
\begin{equation}
    \mathcal{L}_{set} = 
        \alpha_1 \mathcal{L}_{giou}
        + \alpha_2 \mathcal{L}_{cls}
        + \alpha_3 \mathcal{L}_{center-reg} 
        + \alpha_4 \mathcal{L}_{size-reg},
\end{equation}
where $\alpha_1=10$, $\alpha_2=1$, $\alpha_3=5$, $\alpha_4=1$ are set heuristically.
The set loss $\mathcal{L}_{set}$ is applied to all $n_{dec-layer}$ layers in the decoder for better convergence.

\myparagraph{Caption Loss.}
Following the standard practice of image captioning, we train our caption head first with standard cross-entropy loss (MLE training), and then fine-tune it with \textbf{S}elf-\textbf{C}ritical \textbf{S}equence \textbf{T}raining (SCST)\cite{rennie2017scst}. 
During MLE training, the model is trained to predict the $\left(t+1\right)$th word $c_i^{t+1}$, given the first $t$ words $c_i^{[1:t]}$ and the visual clue $\mathcal{V}$.
The loss function for a $T$-length sentence is defined as:
\begin{equation}
    \mathcal{L}_{c_i} = \sum_{i=1}^{T} \mathcal{L}_{c_i}(t) = -\sum_{i=1}^{T} \log \hat{P}\left(c_i^{t+1} \vert \mathcal{V}, c_i^{[1:t]}\right).
\end{equation}
After the caption head is trained under word-level supervision, we fine-tune it with SCST.
During SCST, the model generates multiple captions $\hat{c}_{1, \cdots,k}$ with a beam size of $k$, and another $\hat{g}$ through greedy search as a baseline.
The loss function for SCST is defined as:
\begin{equation}
    \mathcal{L}_{c_i} = 
    - \sum_{i=1}^{k}
        \left(R\left(\hat{c}_{i}\right) - R\left(\hat{g}\right)\right) 
        \cdot
        \frac{1}{\left|\hat{c}_i\right|}\log \hat{P}\left(\hat{c}_i\vert \mathcal{V}\right)
        .
\end{equation}
Here, the reward function $R\left(\cdot\right)$ is the CIDEr metric for caption evaluation, and the log probability of caption $\hat{c}_i$ is normalized by caption length $\left|\hat{c}_i\right|$, to encourage the model to treat captions with different length equally important.

\myparagraph{Set to Set Training for 3D Dense Captioning.}
We propose an easy-to-implement set-to-set training strategy for 3D dense captioning.
Given a 3D scene, we randomly sample one sentence from the corpus for each annotated instance.
Then, we assign language annotations to the corresponding number of proposals in the corresponding scene with the same Hungarian algorithm.
During training, we average losses for captions $\mathcal{L}_{c_i}$ on all annotated instances in a batch, to compute the caption loss $\mathcal{L}_{cap}$.
%
%
%
To balance losses for different tasks, our loss function for the whole system is defined as:
\begin{equation}
    \mathcal{L} = \beta_1\mathcal{L}_{vq} + \beta_2 \sum_{i=1}^{n_{dec-layer}}\mathcal{L}_{set} + \beta_3 \mathcal{L}_{cap},
\end{equation}
where $\beta_1 = 10$, $\beta_2 = 1$, $\beta_3 = 5$ are set heuristically.
\section{Experiments}
\label{sec:experiments}
We first present the datasets, metrics, and implementation details for 3D dense captioning (section \ref{subsec:datasets,metric,implementation}).
Then, we provide comparisons with all state-of-the-art methods (section \ref{subsec:comparison with existing}).
We also provide studies on the effectiveness of different parts in our model (section \ref{subsec:ablation}).
Finally, we visualize several qualitative results to address the effectiveness of our method (section \ref{subsec:qualitative}).

\subsection{Datasets, Metrics, and Implementation Details}
\label{subsec:datasets,metric,implementation}

\myparagraph{Datasets}.
We report results on two commonly used datasets, ScanRefer \cite{chen2020scanrefer} and Nr3D\cite{achlioptas2020referit3d}, both of which are built on 3D scenes from ScanNet\cite{dai2017scannet}.
ScanNet\cite{dai2017scannet} contains 1,201 indoor 3D scenes for training and 312 for validation.
ScanRefer/Nr3D contains 36,665/32,919 free-form language annotations describing 7,875/4,664 objects from 562/511 3D scenes for training, and evaluates on 9,508/8,584 sentences for 2,068/1,214 objects from 141/130 3D scenes.

\myparagraph{Evaluation Metrics}.
Following \cite{chen2021scan2cap,cai20223djcg,jiao2022more,wang2022spacap3d}, we first apply NMS on object proposals to drop duplicate object predictions.
Each object proposal is a box-sentence pair $(\hat{b}_i, \hat{c}_i)$, containing box corner prediction $\hat{b}_i$ and generated sentence $\hat{c}_i$.
Then, each instance is assigned an object proposal with the largest IoU among the remaining proposals.
Here, we use $(b_i, C_i)$ to represent an instance's label, where $b_i$ is a box corner's label and $C_i$ is the corpus containing all caption annotations for this instance.
To jointly evaluate the model's localization and caption generation capability, we adopt the $m@kIoU$ metric\cite{chen2021scan2cap}:
\begin{equation}
    m@kIoU=\frac{1}{N}\sum_{i=1}^{N} m\left(\hat{c}_i, C_i\right) \cdot \mathbb{I}\left\{IoU\left(\hat{b}_i, b_i\right) \ge k\right\}.
\label{eq:m@kIoU}
\end{equation}
Here, $N$ is the number of total annotated instances in the evaluation dataset, and $m$ could be any metric for natural language generation, such as CIDEr\cite{vedantam2015cider}, METEOR\cite{banerjee2005meteor}, BLEU-4\cite{papineni2002bleu}, and ROUGE-L\cite{lin2004rouge}.

\myparagraph{Implementation Details}.
We offer implementation details of different baselines. 
``w/o additional 2D'' means the input $\mathcal{PC}\in \mathbb{R}^{40,000 \times 10}$ contains absolute location as well as \textit{color}, \textit{normal} and \textit{height} for $40,000$ points representing a 3D scene.
``additional 2D'' means we replace color information with $128$-dimensional \textit{multiview} feature extracted by ENet\cite{chen2020hgnet} from 2D images following \cite{chen2021scan2cap}.

%
We first pre-train the whole network without the caption head, on ScanNet\cite{dai2017scannet} detection dataset with ScanRefer\cite{chen2020scanrefer} categories for $1,080$ epochs (about 163k iterations, 34 hours), using the AdamW optimizer\cite{loshchilov2017AdamW} with a learning rate decaying from $5\times 10^{-4}$ to $10^{-6}$ by a cosine annealing scheduler, a weight decay of $0.1$, a gradient clipping of $0.1$, and a batch size of $8$ following \cite{misra2021-3detr}.
Then, we load the pre-trained detector, and train our caption head with MLE loss for another 720 epochs (51k/46k iterations for ScanRefer/Nr3D, 11/10 hours).
To prevent overfitting, we fix the learning rate of the detector as $10^{-6}$, and set that of the caption head decaying from $10^{-4}$ to $10^{-6}$ using another cosine annealing scheduler.
Due to the high memory cost of SCST, we tune the caption head with a batch size of 2 and freeze the detector for 180 epochs (50k/46k iterations for ScanRefer/Nr3D, 14/11 hours) with a fixed learning rate of $10^{-6}$.
We evaluate the model every $2,000$ iterations during training for consistency with existing works\cite{chen2021scan2cap,wang2022spacap3d}, and all experiments mentioned above are conducted on a single RTX3090 GPU.

\subsection{Comparison with Existing Methods}
\label{subsec:comparison with existing}
\begin{table*}[htbp]
    \centering
    \resizebox{\linewidth}{!}{
    \begin{tabular}{cccccccccccccccccccccc}
    \toprule
    \multirow{3}{*}{Method} & \multirow{3}{*}{$\mathcal{L}_{des}$} &  & \multicolumn{9}{c}{w/o additional 2D input}                                                                          &  & \multicolumn{9}{c}{w/ additional 2D input}                                                                            \\
                            &                                      &  & \multicolumn{4}{c}{IoU = 0.25}                          &  & \multicolumn{4}{c}{IoU = 0.50}                          &  & \multicolumn{4}{c}{IoU = 0.25}                          &  & \multicolumn{4}{c}{IoU = 0.50}                          \\ \cline{4-7} \cline{9-12} \cline{14-17} \cline{19-22} 
                            &                                      &  & C$\uparrow$ & B-4$\uparrow$ & M$\uparrow$ & R$\uparrow$ &  & C$\uparrow$ & B-4$\uparrow$ & M$\uparrow$ & R$\uparrow$ &  & C$\uparrow$ & B-4$\uparrow$ & M$\uparrow$ & R$\uparrow$ &  & C$\uparrow$ & B-4$\uparrow$ & M$\uparrow$ & R$\uparrow$ \\ \hline
    Scan2Cap\cite{chen2021scan2cap}                & \multirow{7}{*}{MLE}                 &  & 53.73       & 34.25         & 26.14       & 54.95       &  & 35.20       & 22.36         & 21.44       & 43.57       &  & 56.82       & 34.18         & 26.29       & 55.27       &  & 39.08       & 23.32         & 21.97       & 44.78       \\
    MORE\cite{jiao2022more}                    &                                      &  & 58.89       & 35.41         & 26.36       & 55.41       &  & 38.98       & 23.01         & 21.65       & 44.33       &  & 62.91       & 36.25         & 26.75       & 56.33       &  & 40.94       & 22.93         & 21.66       & 44.42       \\
    SpaCap3d\cite{wang2022spacap3d}   &                                      &  & 58.06       & 35.30         & 26.16       & 55.03       &  & 42.76       & 25.38         & 22.84       & 45.66       &  & 63.30       & 36.46         & 26.71       & 55.71       &  & 44.02       & 25.26         & 22.33       & 45.36       \\
    3DJCG\cite{cai20223djcg}                   &                                      &  & 60.86       & \textbf{39.67}         & 27.45       & 59.02       &  & 47.68       & 31.53         & 24.28       & 51.80       &  & 64.70       & \textbf{40.17}     & 27.66       & \textbf{59.23}       &  & 49.48       & 31.03         & 24.22       & 50.80       \\
    D3Net\cite{chen2021d3net}         &                                      &  & -           & -             & -           & -           &  & -           & -             & -           & -           &  & -           & -             & -           & -           &  & 46.07       & 30.29         & 24.35       & 51.67       \\
    Ours                    &                                      &  & \textbf{71.45}       & 39.34         & \textbf{28.25}       & \textbf{59.33}                                                 &  & \textbf{61.81}       & \textbf{34.46}         & \textbf{26.22}       & \textbf{54.40}                                                 &  & \textbf{72.79}       & 39.17         & \textbf{28.06}       & \textbf{59.23}                                                 &  &      \textbf{59.32}       & \textbf{32.42}         & \textbf{25.28}            & \textbf{52.53}       \\ \hline
    $\chi$-Trans2Cap\cite{yuan2022x-trans2cap}        & \multirow{5}{*}{SCST}                &  & 58.81       & 34.17         & 25.81       & 54.10       &  & 41.52       & 23.83         & 21.90       & 44.97       &  & 61.83       & 35.65         & 26.61       & 54.70       &  & 43.87       & 25.05         & 22.46       & 45.28       \\
    Scan2Cap\cite{chen2021scan2cap}                &                                      &  & -           & -             & -           & -           &  & -             & -             & -           & -           &  & -           & -             & -           & -           &  & 48.38       & 26.09         & 22.15       & 44.74       \\
    D3Net\cite{chen2021d3net}                   &                                      &  & -           & -             & -           & -           &  & -           & -           & -           & -           &  & -           & -             & -           & -           &  & 62.64       & 35.68         & \textbf{25.72}       & \textbf{53.90}       \\
    Ours                    &                                      &  & \textbf{84.15}       & \textbf{42.51}         & \textbf{28.47}       & \textbf{59.26}       &  & \textbf{73.77}       & \textbf{38.21}       & \textbf{26.64}       & \textbf{54.71}         &  & \textbf{86.28} & \textbf{42.64}       &  \textbf{28.27}    & \textbf{59.07}   &  & \textbf{70.63}    & \textbf{35.69}     & 25.51   & 52.28             \\ \bottomrule
    \end{tabular}
    }
    \setlength{\abovecaptionskip}{0.2cm}
    \caption{
    \textbf{Evaluating Vote2Cap-DETR on ScanRefer\cite{chen2020scanrefer}.}
    We compare Vote2Cap-DETR with all published state-of-the-art 3D dense caption methods on the ScanRefer dataset.
    Though our method does not depend on hand-crafted NMS\cite{neubeck2006nms} to drop overlapped boxes, we follow the standard evaluation protocol from \cite{chen2021scan2cap} for fair comparison and provide evaluation without NMS in Table \ref{tab:effect-nms}.
    Our proposed Vote2Cap-DETR achieves new state-of-the-art under both MLE training and SCST.
    }
    \label{exp:comparison_on_scanrefer}
\end{table*}
In this section, we compare performance with existing works on metrics \textbf{C}, \textbf{M}, \textbf{B-4}, \textbf{R} as abbreviations for CIDEr\cite{vedantam2015cider}, METEOR\cite{banerjee2005meteor}, BLEU-4\cite{papineni2002bleu}, Rouge-L\cite{lin2004rouge} under IoU thresholds of 0.25, 0.5 for ScanRefer (Table \ref{exp:comparison_on_scanrefer}) and 0.5 for Nr3D (Table \ref{exp:comparison_on_nr3d}).
``-'' indicates that neither the original paper nor any follow-up works provide such results.
Since different supervision on the caption head has a huge influence on the captioning performance, we make separate comparisons for MLE training and SCST.
%
%
Among all the listed methods, experiments other than D3Net\cite{chen2021d3net} and 3DJCG\cite{cai20223djcg} utilize the standard VoteNet\cite{qi2019votenet} detector.
Meanwhile, D3Net\cite{chen2021d3net} adopts PointGroup\cite{jiang2020pointgroup}, a 3D instance segmentation model, for better object detection.
3DJCG\cite{cai20223djcg} improves VoteNet's localization performance with an FCOS\cite{tian2019fcos} head, which predicts distance from a voting point to each side of a bounding box.
%
%
Additionally, 3DJCG and D3Net focus on the joint promotion of 3D dense captioning and 3D visual grounding, therefore their reported models are trained with data from both tasks.
%
%
Among methods listed under SCST, $\chi$-Trans2Cap\cite{yuan2022x-trans2cap} combines MLE training with standard SCST in an additive manner, Scan2Cap and D3Net\cite{chen2021d3net} adopt the same reward combining CIDEr score and listener losses with a weighted sum.
It's worth mentioning that our model adopts the standard SCST, whose reward function is CIDEr score.

Table \ref{exp:comparison_on_scanrefer} reports comparisons on ScanRefer\cite{chen2020scanrefer} validation dataset.
Our Vote2Cap-DETR surpasses current state-of-the-art methods.
For example, under MLE training with additional 2D inputs, our Vote2Cap-DETR achieves 59.32\% C@0.5 while 3DJCG\cite{cai20223djcg} achieves 49.48\% (9.84\% C@0.5$\uparrow$) with additional training data.
Additionally, under SCST, our Vote2Cap-DETR achieves 70.63\% C@0.5, while 62.64\% (7.99\% C@0.5$\uparrow$) for current state-of-the-art D3Net\cite{chen2021d3net} with more training labels and semi-supervised training on more training data.

In Table \ref{exp:comparison_on_nr3d}, we list results on the Nr3D\cite{achlioptas2020referit3d} dataset with additional 2D input following \cite{wang2022spacap3d}.
Since Scan2Cap\cite{chen2021scan2cap} has not reported results on Nr3D, we adopt the best-reported result from \cite{cai20223djcg}.
Our proposed Vote2Cap-DETR also surpasses current state-of-the-art methods.

\begin{table}[htbp]
    \centering
    \resizebox{\linewidth}{!}{
    \begin{tabular}{cccccc}
    \toprule
    Method          & $\mathcal{L}_{des}$   & C@0.5$\uparrow$ & B-4@0.5$\uparrow$ & M@0.5$\uparrow$ & R@0.5$\uparrow$ \\ \hline
    Scan2Cap\cite{chen2021scan2cap}        & \multirow{6}{*}{MLE}  & 27.47           & 17.24             & 21.80           & 49.06           \\
    SpaCap3d\cite{wang2022spacap3d}        &                       & 33.71           & 19.92             & 22.61           & 50.50           \\
    D3Net\cite{chen2021d3net}           &                       & 33.85           & 20.70             & 23.13           & 53.38           \\
    3DJCG\cite{cai20223djcg}           &                       & 38.06           & 22.82             & 23.77           & 52.99           \\
    Ours            &                       & \textbf{43.84}  & \textbf{26.68}    & \textbf{25.41}  & \textbf{54.43}  \\ \hline
    $\chi$-Tran2Cap\cite{yuan2022x-trans2cap} & \multirow{4}{*}{SCST} & 33.62           & 19.29             & 22.27           & 50.00           \\
    D3Net\cite{chen2021d3net}           &                       & 38.42           & 22.22             & 24.74           & 54.37           \\
    Ours            &                       & \textbf{45.53}  & \textbf{26.88}    & \textbf{25.43}  & \textbf{54.76}  \\ \bottomrule
    \end{tabular}
    }
    \setlength{\abovecaptionskip}{0.2cm}
    \caption{
    \textbf{Evaluating Vote2Cap-DETR on Nr3D\cite{achlioptas2020referit3d}.}
    Likewise, we perform the standard evaluation on the Nr3D dataset, and our proposed Vote2Cap-DETR surpasses prior arts.
    }
    \label{exp:comparison_on_nr3d}
\end{table}

\subsection{Ablation Study}
\label{subsec:ablation}
Since 3D dense captioning concerns both localization and caption generation, we perform ablation studies to understand the effectiveness of different components.

\begin{figure}[htbp]
	\centering
	\includegraphics[width=\linewidth]{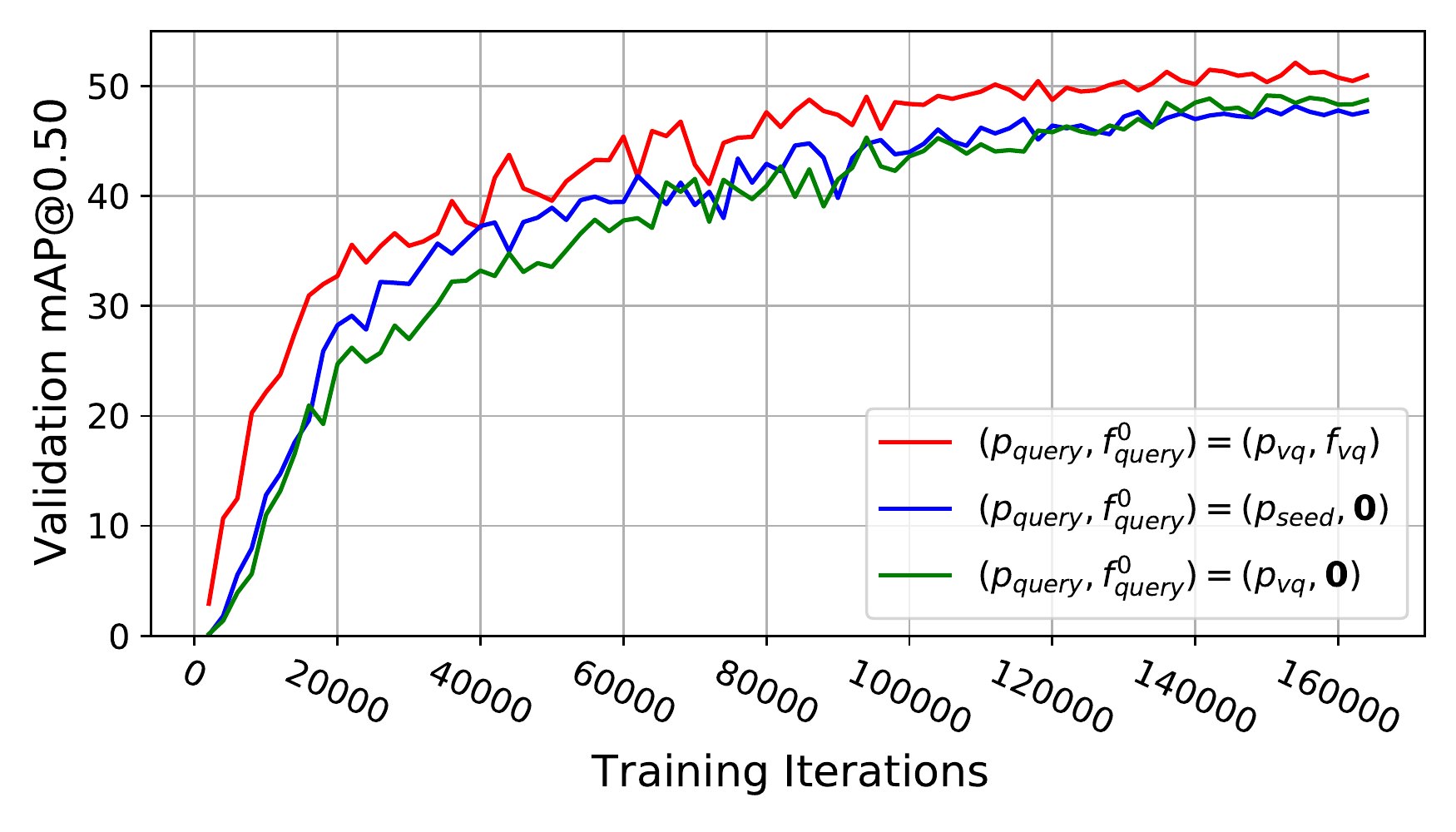}
	\caption{
	\textbf{Vote query and convergence.}
	We take out convergence study on a different combination of content feature $f_{vq}$ and position $p_{vq}$ in vote query.
	The baseline model $(p_{query}, f^0_{query}) = (p_{seed}, \mathbf{0})$ downgrades to 3DETR.
	Introducing $p_{vq}$ boosts performance but decelerates training since $FFN_{vote}$ requires time to converge, and $f_{vq}$ accelerates training.
	%
	}
	\label{fig:detr-converge}
\end{figure}

\myparagraph{Does the vote query improve 3DETR?} We performed ablation experiments in Table \ref{tab:ablation_detection} and Figure \ref{fig:detr-converge}  to see if the vote query can improve 3DETR's localization and convergence. 
%
%
%
Introducing position features $p_{vq}$ alone helps improve detection performance (0.97\% mAP50$\uparrow$). 
%
%
However, it (green line in Figure \ref{fig:detr-converge}) converges slower in the earlier training procedure than the 3DETR baseline (blue line in Figure \ref{fig:detr-converge}), inferring the vote query generation module is not well learned to predict accurate spatial offset estimations at early training epochs.
Introducing additional content feature $f_{vq}$ in vote query features 
results in another boost in both detection performance (2.98\% mAP50$\uparrow$) and training speed (red line in Figure \ref{fig:detr-converge}).
%
The overall localization performance of Vote2Cap-DETR is about 7.2\% mAP higher than the popular VoteNet.
%
%

\begin{table}[htb]
    \centering
    \resizebox{\linewidth}{!}{
    \begin{tabular}{ccccccccc}
    \hline
    \multirow{2}{*}{$p_{query}$} & \multirow{2}{*}{$f^0_{query}$} & \multicolumn{2}{c}{IoU=0.25} & \multicolumn{2}{c}{IoU=0.50} &  & \multicolumn{2}{c}{$1$st layer IoU=0.50} \\ \cline{3-6} \cline{8-9} 
                                 &                                & mAP$\uparrow$ & AR$\uparrow$ & mAP$\uparrow$ & AR$\uparrow$ &  & mAP$\uparrow$       & AR$\uparrow$       \\ \cline{1-6} \cline{8-9} 
    \multicolumn{2}{c}{VoteNet Baseline}                          & 63.42         & 82.18        & 44.96         & 60.65        &  & -                   & -                  \\ \cline{1-2}
    $p_{seed}$                   & $\mathbf{0}$                   & 67.25         & 84.91        & 48.18         & 64.98        &  & 34.80          & 55.06              \\
    $p_{vq}$                     & $\mathbf{0}$                   & 67.33         & 85.60        & 49.15         & 66.38        &  & 30.23          & 58.44              \\
    $p_{vq}$                     & $f_{vq}$                       & \textbf{69.61} & \textbf{87.20} & \textbf{52.13}   & \textbf{69.12}       &  & 46.53               & 66.51          \\ \hline
    \end{tabular}
    }
    \setlength{\abovecaptionskip}{0.2cm}
    \caption{
        \textbf{Vote query and performance.} We provide quantitative results for Figure \ref{fig:detr-converge}. 
        Introducing $p_{vq}$ as query positions improves detection, and gathering $f_{vq}$ from content further boosts performance.
    }
    \label{tab:ablation_detection}
\end{table}

\myparagraph{Does 3D context feature help captioning?} Since the performance of 3D dense captioning is affected by both localization and caption capability, we freeze all parameters other than the caption head, and train with 3D only input and standard cross entropy loss (MLE training) for a fair evaluation.
We use object-centric decoder\cite{wang2022spacap3d} as our baseline, which is a decoder that generates captions with object feature as a caption's prefix.
%
%
In Table \ref{tab:ablation_caption_memory}, 
``-'' refers to the object-centric decoder baseline, 
``global'' means naively including all context tokens extracted from the scene encoder in the decoder,
%
``local'' is our proposed caption head that includes a vote query's $k_s$ ($k_s=128$ empirically) nearest context tokens extracted from the scene encoder.

With the object feature as a caption's prefix, caption generation performance benefits from introducing additional contextual information.
%
%
Additionally, compared with naively introducing contextual information from the whole scene, introducing local information could be more beneficial.
This demonstrates our motivation that close surroundings matter when describing an object.
\begin{table}[htbp]
    \resizebox{\linewidth}{!}{
    \begin{tabular}{ccccccccccc}
    \toprule
    \multirow{2}{*}{key} &  & \multicolumn{4}{c}{IoU=0.25}                                      &  & \multicolumn{4}{c}{IoU=0.5}                                       \\ \cline{3-6} \cline{8-11} 
                         &  & C$\uparrow$    & B-4$\uparrow$  & M$\uparrow$    & R$\uparrow$    &  & C$\uparrow$    & B-4$\uparrow$  & M$\uparrow$    & R$\uparrow$    \\ \cline{1-1} \cline{3-6} \cline{8-11} 
    -                    &  & 68.62          & 38.61          & 27.67          & 58.47          &  & 60.15          & 34.02          & 25.80          & 53.82          \\
    global               &  & 70.05          & 39.23          & 27.84          & 58.44          &  & 61.20          & 34.66          & 25.93          & 53.79          \\
    local                &  & \textbf{70.42} & \textbf{39.98} & \textbf{27.99} & \textbf{58.89} &  & \textbf{61.39} & \textbf{35.24} & \textbf{26.02} & \textbf{54.12} \\ \bottomrule
    \end{tabular}
    }
    \setlength{\abovecaptionskip}{0.2cm}
    \caption{
    \textbf{Different keys for caption generation.}
    We provide a comparison on different keys used in caption generation.
    Introducing contextual information relates to more informative captions generated.
    Since 3D dense captioning is more object-centric, introducing vote queries' local contextual feature is a better choice.
    }
    \label{tab:ablation_caption_memory}
\end{table}

\begin{figure*}[htbp]
	\centering
	\includegraphics[width=\linewidth]{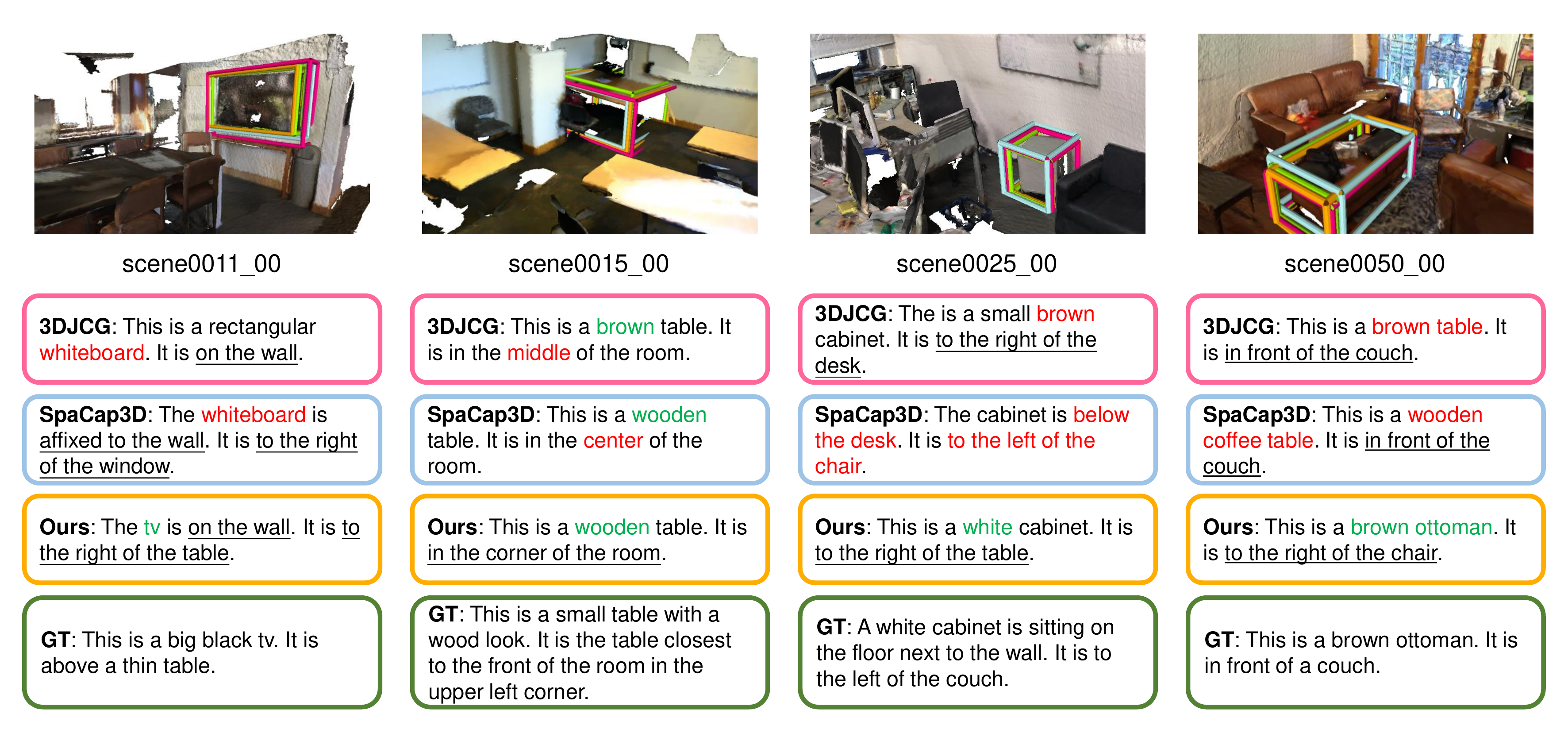}
	\caption{
	\textbf{Qualitative Comparisons.}
	We compare qualitative results with two state-of-the-art ``detect-then-describe'' methods, 3DJCG\cite{cai20223djcg} and SpaCap3D\cite{wang2022spacap3d}.
	We \underline{underline} phrases describing spatial locations, and mark correct attribute words in \textcolor{green}{green} and wrong descriptions in \textcolor{red}{red}.
	Our method produces tight bounding boxes close to ground truth annotations and produce accurate descriptions of object attributes, classes and spatial relationship.
	}
	\label{fig:qualitative_comparison}
\end{figure*}

\begin{figure}[htbp]
	\centering
	\includegraphics[width=\linewidth]{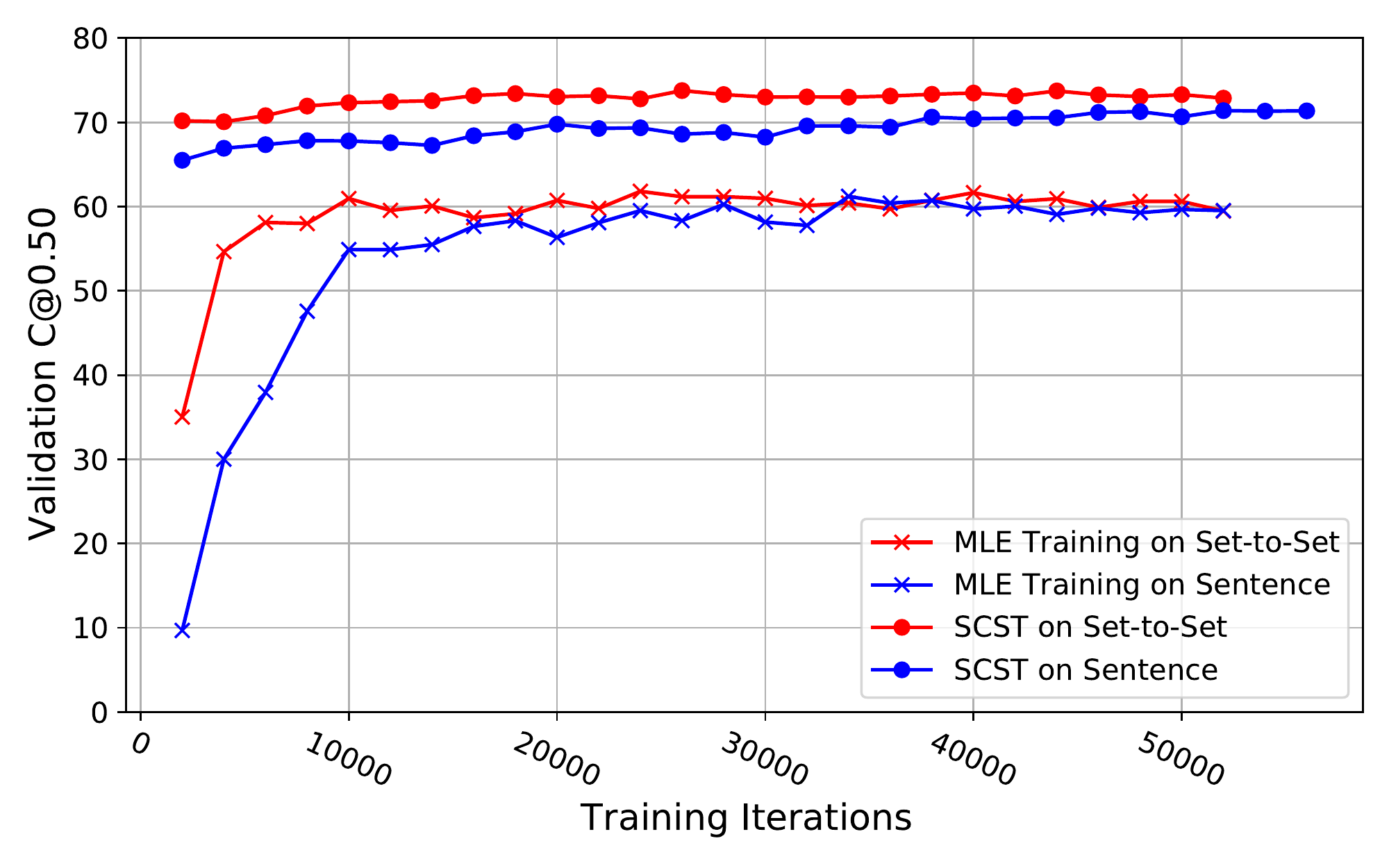}
	\caption{
	\textbf{Set-to-Set training and convergence.} 
	Convergence speed analysis of two different training strategies with MLE training as well as SCST.
	Set-to-Set training enables a larger batch size for the caption head, which accelerates convergence on 3D dense captioning.
	}
	\label{fig:set-to-set}
\end{figure}

\myparagraph{Do Set-to-Set Training benefit dense captioning?}
To analyze effectiveness of set-to-set training, we follow the training procedure that utilize a smaller learning rate for all parameters other than the caption head, and freeze these parameters during SCST.
We name the baseline training strategy as ``Sentence Training'', which traverses through all sentence annotations in the dataset and is widely adopted in various works\cite{chen2021scan2cap,wang2022spacap3d}.
As is shown in Figure \ref{fig:set-to-set}, our proposed ``Set-to-Set'' training achieves comparable results with the traditional ``Sentence Training'' during MLE training, and converges faster because of a bigger batch size on the caption head, which also benefits SCST.
\begin{table}[htbp]
    \centering
    \resizebox{\linewidth}{!}{
    \begin{tabular}{cccccc}
    \toprule
    Training                 & $\mathcal{L}_{des}$   & C@0.5$\uparrow$ & B-4@0.5$\uparrow$ & M@0.5$\uparrow$ & R@0.5$\uparrow$ \\ \hline
    Sentence                 & \multirow{2}{*}{MLE}  & 61.21           & \textbf{35.35}    & 26.12           & \textbf{54.52}  \\
    Set-to-Set               &                       & \textbf{61.81}  & 34.46             & \textbf{26.22}  & 54.40           \\ \hline
    Sentence                 & \multirow{2}{*}{SCST} & 71.39           & 37.57             & 26.01           & 54.28           \\
    Set-to-Set               &                       & \textbf{73.77}  & \textbf{38.21}           & \textbf{26.64}        & \textbf{54.71}  \\ \bottomrule
    \end{tabular}
    }
    \setlength{\abovecaptionskip}{0.2cm}
    \caption{
    \textbf{Set to Set training and performance.} 
    We compare our proposed set-to-set training with traditional ``Sentence Training'', which traverses through all sentence annotations.
    We achieve comparable performance with MLE training, and 2.38\% C@0.5 improvement with SCST.
    }
    \label{tab:set-to-set-training}
\end{table}

\myparagraph{Is Vote2Cap-DETR robust to NMS?}
%
Similar to other DETR works, the set loss will encourage the model to produce compact predictions.
We compare performance on both 3D dense caption (C@0.5) and detection (mAP50, AR50) in Table \ref{tab:effect-nms}.
Since the $m@kIoU$ metric (Eq. \ref{eq:m@kIoU}) does not contain any penalties on redundant predictions, getting rid of NMS\cite{neubeck2006nms} results in performance growth on C@0.5.
Absence of NMS restricts the detection precision performance (mAP50) of SpaCap3D (14.47\% mAP50 $\downarrow$) and 3DJCG (17.55\% mAP50 $\downarrow$), however that of Vote2Cap-DETR remains stable.
\begin{table}[htbp]
    \centering
    \resizebox{\linewidth}{!}{
    \begin{tabular}{ccccccccc}
    \toprule
    \multirow{2}{*}{Models} & \multicolumn{3}{c}{w/ NMS}                              & \multicolumn{3}{c}{w/o NMS}                            \\ \cline{2-7} 
                            & C@0.5$\uparrow$ & mAP50$\uparrow$ & AR50$\uparrow$ & C@0.5$\uparrow$ & mAP50$\uparrow$ & AR50$\uparrow$ \\ \hline
    SpaCap3D                & 43.93           & 37.77             & 53.96            & 51.35           & 23.30             & 64.14            \\
    3DJCG                   & 50.22           & 47.58             & 62.12            & 54.94           & 30.03             & 68.69            \\
    Vote2Cap-DETR           & 70.63           & 52.79             & 66.09            & 71.57           & 52.82             & 67.80            \\ \bottomrule
    \end{tabular}
    }
    \setlength{\abovecaptionskip}{0.2cm}
    \caption{
    \textbf{Effect of NMS.} We analyze whether the absence of NMS affects the 3D dense captioning performance (C@0.5) as well as detection performance (mAP50, AR50).
    }
    \label{tab:effect-nms}
\end{table}

\subsection{Qualitative Results}
\label{subsec:qualitative}
We compare qualitative results with two state-of-the-art models, SpaCap3D\cite{wang2022spacap3d} and 3DJCG\cite{cai20223djcg} in Figure\ref{fig:qualitative_comparison}.
One can see that our method produces tight bounding boxes close to the ground-truth. Moreover, our method can produce accurate descriptions of object attributes, classes, and spatial relationships. 
\section{Conclusion.}
In this work,  we present Vote2Cap-DETR, a transformer based one-stage approach, for 3D dense captioning.
%
The proposed Vote2Cap-DETR adopts a fully transformer encoder-decoder architecture that decodes a set of vote queries to box predictions and captions in parallel.
We show that by introducing spatial bias and content-aware features, vote query boosts both convergence and detection performance.
Additionally, we develop a novel lightweight query-driven caption head for informative caption generation.
%
Experiments on two widely used datasets for 3D dense captioning validates that our propose one-stage Vote2Cap-DETR model surpasses prior works with heavy dependence on hand-crafted components by a large margin.

{\small
\bibliographystyle{ieee_fullname}
\bibliography{reference}
}

\clearpage
\onecolumn
\section*{\hfil {\LARGE Appendix}\hfil}
\vspace{50pt}
\renewcommand\thesection{\Alph{section}}
\setcounter{section}{0}
%
In our supplementary material, we first propose a non-transformer baseline for our method that builds on VoteNet\cite{qi2019votenet} in section \ref{supp:votenet-one-stage}.
%
Then, we provide additional experimental details in section \ref{supp:additional-experiments}.
%
Finally, we provide several qualitative studies in section \ref{supp:visualization}.
It's also worth mentioning that our proposed Vote2Cap-DETR sets \textbf{a new state-of-the-art} on the Scan2Cap online test benchmark (Figure \ref{fig:test-benchmark}).


\section{VoteNet baseline with set-to-set training}
\label{supp:votenet-one-stage}

%
In this section, we perform ablation study by replacing our Vote2Cap-DETR's components (SceneEncoder, Vote Query, Transformer Decoder) with VoteNet to study the behavior of non-transformer architecture's behavior.
%
%
%
%
In Table \ref{tab:votenet-one-stage-dense-caption}, we observe that without delicate hand-crafted relation modelling modules, the VoteNet baseline surpasses 3DJCG\cite{cai20223djcg} by 3.48 in C@0.5 $\uparrow$ and 6.23 in C@0.25 $\uparrow$ and achieves comparable results on other metrics with MLE training. 
The results demonstrate the novel caption head and set-to-set training can also improve non-transformer architecture's dense captioning performance. 
%
On the other hand, the VoteNet baseline still falls short in terms of our Vote2Cap-DETR, which demonstrates that Vote Query can help learn more discriminate features in an end-to-end manner for end tasks without resorting to many hand-crafted components as in VoteNet.

\begin{table*}[htbp]
    \centering
    \resizebox{0.7\linewidth}{!}{
    \begin{tabular}{ccccccccccc}
    \toprule
    \multirow{2}{*}{Method} &  & \multicolumn{4}{c}{IoU = 0.25}                          &  & \multicolumn{4}{c}{IoU = 0.5}                           \\ \cline{3-6} \cline{8-11}
                    &  & C$\uparrow$ & B-4$\uparrow$ & M$\uparrow$ & R$\uparrow$ &  & C$\uparrow$ & B-4$\uparrow$ & M$\uparrow$ & R$\uparrow$ \\ \cline{1-1} \cline{3-6} \cline{8-11} 
    3DJCG\cite{cai20223djcg}                   &  & 64.70       & \textbf{40.17}& 27.66       & \textbf{59.23}       &  & 49.48       & \textbf{31.03}         & 24.22       & \textbf{50.80}       \\
    Ours(VoteNet)           &  & \textbf{70.93}& 39.92         & \textbf{28.09}& 58.88       &  & \textbf{52.96}       & 30.59         & \textbf{24.40}       & 50.10       \\\hline
    Ours(Full)              &  & \textbf{72.79}       & 39.17         & 28.06          & \textbf{59.23}                                                 &  &      \textbf{59.32}       & \textbf{32.42}         & \textbf{25.28}            & \textbf{52.53}       \\
    \bottomrule
    \end{tabular}
    }
    \vspace{2mm}
    \caption{
    \textbf{VoteNet baseline with set-to-set training.} We replace Vote2Cap-DETR's components with VoteNet. One can see that non-transformer VoteNet architecture also benefits from our novel caption head and set to set training. Although there are still performance gaps with our Vote2Cap-DETR architecture.  
    }
    \label{tab:votenet-one-stage-dense-caption}
\end{table*}

\section{Experiments}
\label{supp:additional-experiments}

We provide evaluations on the Scan2Cap online test benchmark (section \ref{supp:scanrefer test}) as well as additional experimental details (section \ref{supp:per-class results} \& \ref{supp:implementation-details}) in this section.

\subsection{Scan2Cap Test Benchmark}
\label{supp:scanrefer test}
\begin{figure*}[htbp]
	\centering
	\includegraphics[width=\linewidth]{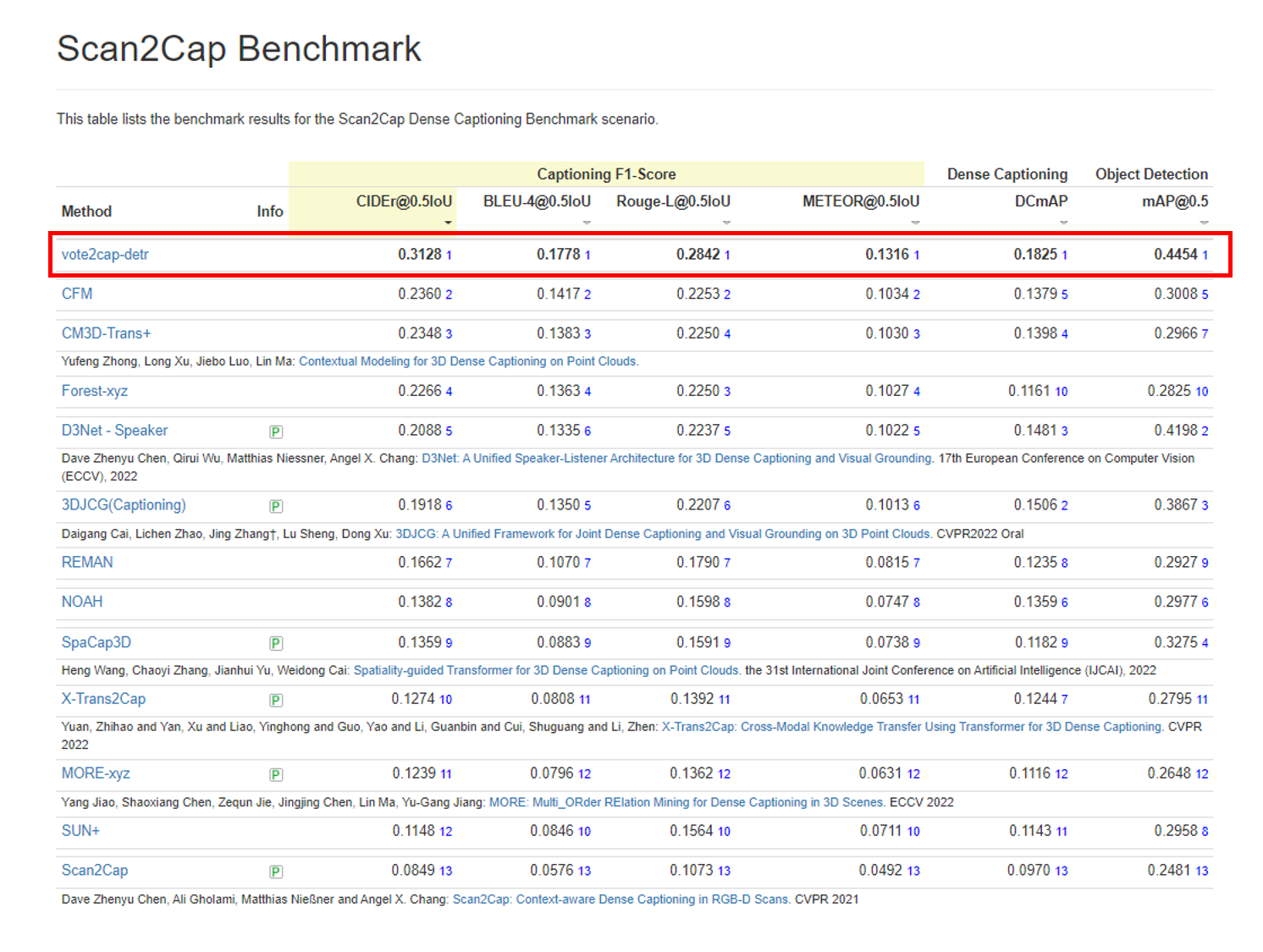}
	\caption{
	\textbf{Scan2Cap\cite{chen2021scan2cap} test benchmark.}
	Our proposed Vote2Cap-DETR achieves a new state-of-the-art for all metrics on the Scan2Cap online test benchmark.
	}
	\label{fig:test-benchmark}
\end{figure*}

Our proposed Vote2Cap-DETR achieves a new state-of-the-art for all metrics on the Scan2Cap online test benchmark (Figure \ref{fig:test-benchmark}, \href{https://kaldir.vc.in.tum.de/scanrefer\_benchmark/benchmark\_captioning}{https://kaldir.vc.in.tum.de/scanrefer\_benchmark/benchmark\_captioning}).

\subsection{Per-Class mAP Results}
\label{supp:per-class results}
We list per class mAP results for VoteNet\cite{qi2019votenet}, 3DETR\cite{misra2021-3detr}, and our proposed Vote2Cap-DETR on ScanNet scenes\cite{dai2017scannet} under an IoU threshold of 0.5 in Table \ref{tab:ScanNet AP per class}.
The overall performance is listed in the main paper.

\begin{table*}[htbp]
    \centering
    \resizebox{\linewidth}{!}{
    \begin{tabular}{ccccccccccccccccccc}
    \toprule
    Method        & cabinet & bed   & chair & sofa  & table & door  & window & bookshelf & picture & counter & desk  & curtain & refrigerator & shower curtain & toilet & sink  & bathtub & others \\ \hline
    VoteNet\cite{qi2019votenet}       & 21.41   & 78.41 & 78.47 & 74.44 & 55.42 & 34.68 & 14.91  & 29.80     & 9.04    & 16.57   & 51.12 & 34.62   & 40.12        & 45.82          & 89.93  & 37.23 & 83.41   & 13.79  \\
    3DETR\cite{misra2021-3detr}         & 26.30   & 75.78 & 82.19 & 59.15 & 62.25 & 39.16 & 21.47  & 33.14     & 16.45   & 34.41   & 49.68 & 38.34   & 42.83        & 33.33          & 88.68  & 52.62 & 82.41   & 29.06  \\
    Vote2Cap-DETR & 31.98   & 81.48 & 85.80 & 64.37 & 65.20 & 41.19 & 28.47  & 39.81     & 22.94   & 39.02   & 54.46 & 36.66   & 40.19        & 56.10          & 87.97  & 44.38 & 85.12   & 33.28  \\ \bottomrule
    \end{tabular}
    }
    \setlength{\abovecaptionskip}{0.2cm}
    \caption{\textbf{Per-class AP under IoU threshold of 0.5 on ScanNet scenes.}}
    \label{tab:ScanNet AP per class}
\end{table*}

\subsection{Implementation Details}
\label{supp:implementation-details}

Our proposed Vote2Cap-DETR first goes through the feature encoding module, then we generate vote queries from the encoded feature as object queries, and we decode the vote queries to bounding boxes and captions in the end.

\textbf{Feature Encoding}
directly operates on the input point cloud $PC$ to 1,024 tokens with a feature size of 256.
We first tokenizes the input point cloud $PC = [p_{in}; f_{in}]\in \mathbb{R}^{40,000\times (3 + d_{in})}$ to point tokens $[p_{token}; f_{token}]\in \mathbb{R}^{2,048 \times (3 + 256)}$ with a set-abstraction layer\cite{qi2017pointnet++} with hidden sizes of [$3 + d_{in}$, 64, 128, 256].
Then, our scene encoder encodes point tokens $[p_{token}; f_{token}] \in \mathbb{R}^{2,048 \times (3 + 256)}$ to $[p_{enc}; f_{enc}] \in \mathbb{R}^{1,024 \times (3 + 256)}$.
We adopt the same encoder as 3DETR-m\cite{misra2021-3detr}, which contains a three-layer transformer encoder with a set-abstraction layer between the first two layers.
Each encoder layer has a feature size of 256 and \textbf{F}eed \textbf{F}orward \textbf{N}etwork (FFN) with a hidden size of 128.
The first encoder layer operates on 2,048 points, while the last two operates on the 1,024 points downsampled by the set-abstraction layer.
Additionally, three binary attention masks are applied to each encoder layer with a radius of [0.16, 0.64, 1.44] respectively to force the interactions of points in a given radius.

\textbf{Vote Query Generator} generates 256 object queries $[p_{vq}; f_{vq}] \in \mathbb{R}^{256 \times (3 + 256)}$ from the encoded points $[p_{enc}; f_{enc}] \in \mathbb{R}^{1,024 \times (3 + 256)}$.
It contains an FFN $FFN_{vote}$ with a hidden size of 256 to generate offset estimation and feature projection with respective to $f_{enc}$. 
It also use a set abstraction layer to gather feature $f_{vq} \in \mathbb{R}^{256 \times 256}$ from encoded scene feature for $p_{vq} \in \mathbb{R}^{256 \times 3}$ as described in the main paper.

\textbf{Parallel Decoding} aims to decode the vote queries $[p_{vq}; f_{vq}]$ to corresponding box estimations and captions.
The transformer decoder consists of eight identical transformer decoder layers with four heads for both self-attention and cross-attention.
It operates on vote queries $[p_{vq}; f_{vq}]$ and encoded feature $[p_{enc}; f_{enc}]$ for the final query feature $[p_{vq}, f_{out}] \in \mathbb{R}^{256 \times (3 + 256)}$.
Follow the transformer decoder are two parallel heads, the detection head and the caption head.
The detection head generates center offset estimation ($[-0.5, 0.5]^3$) from vote queries' absolute location $p_{vq}$, normalized size estimation ($[0, 1]^3$), and semantic class estimation from $f_{out}$ using separate FFN heads with a hidden size of 256.
Note that we do not estimate the rotation angles since ScanNet\cite{dai2017scannet} does not contain any rotated boxes.
Our proposed caption head, DCC, generates captions with respect to final query features $f_{out}$ as $\mathcal{V}^{q}$ and $p_{vq}$'s surrounding contextual features $\mathcal{V}^{s}$.
DCC is a two layer transformer decoder with four heads for multi-head attentions, as well as a feature size of 256, a sinusoid position encoding, and a vocabulary of 3,433 for ScanRefer\cite{chen2020scanrefer} and 2,937 for Nr3D\cite{achlioptas2020referit3d}.

\section{Qualitative Results}
\label{supp:visualization}

\begin{figure*}[htbp]
	\centering
	\includegraphics[width=0.98\linewidth]{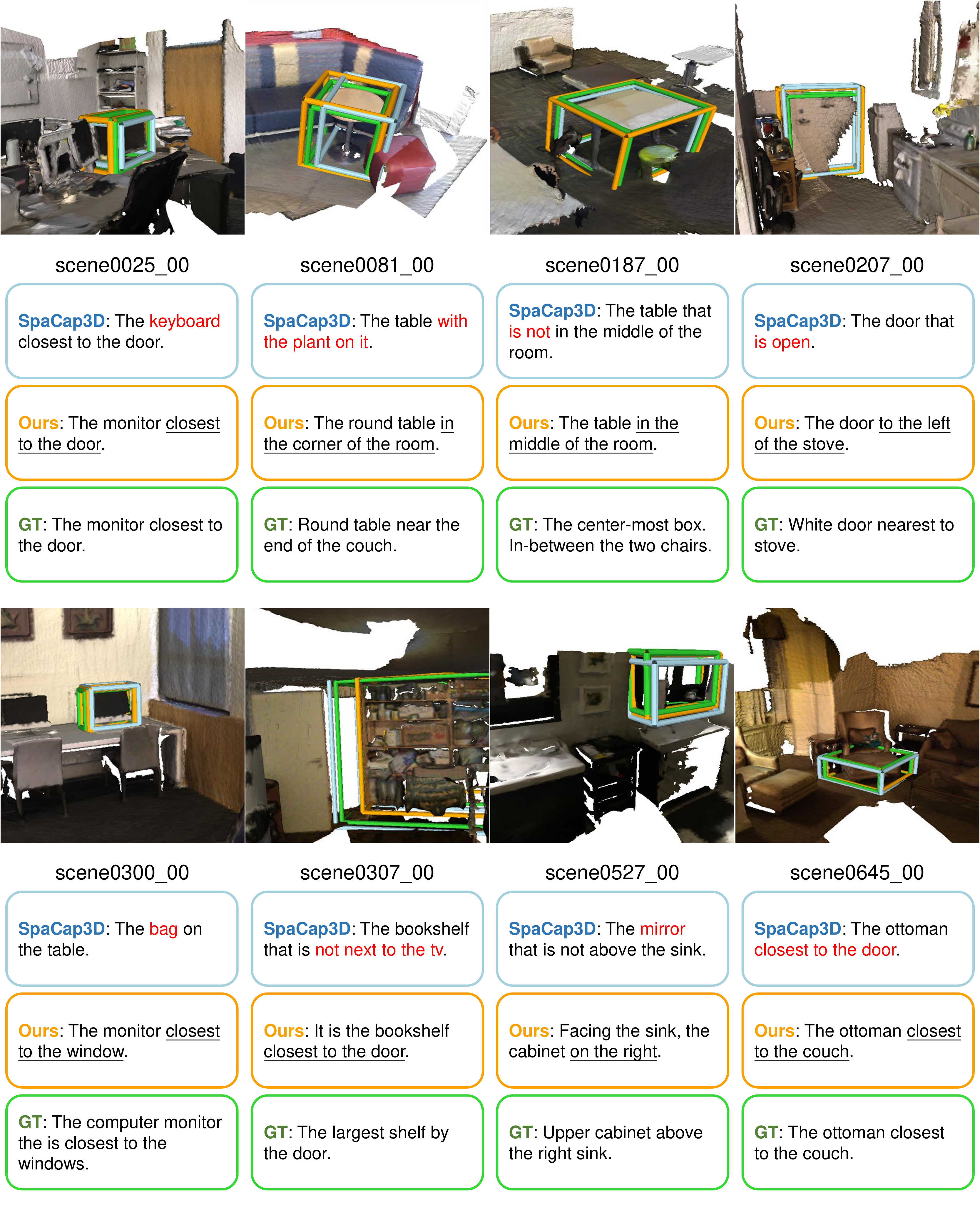}
	\caption{
	\textbf{Visualization of 3D dense captioning on Nr3D\cite{achlioptas2020referit3d}.}
	We visualize several results generated by our proposed Vote2Cap-DETR comparing with SpaCap3D\cite{wang2022spacap3d} on the Nr3D\cite{wang2022detr3d} dataset.
	Our proposed method generates tight bounding box as well as accurate descriptions.
	}
	\label{fig:visualization-nr3d}
\end{figure*}

\textbf{Qualitative results on Nr3D.}
We showcase qualitative results on 3D dense captioning on the Nr3D\cite{achlioptas2020referit3d} dataset in Figure \ref{fig:visualization-nr3d}.
Our proposed Vote2Cap-DETR is also able to generate tight bounding boxes as well as accurate descriptions for each object in a 3D scene.

\begin{figure*}[htbp]
	\centering
	\includegraphics[width=0.98\linewidth]{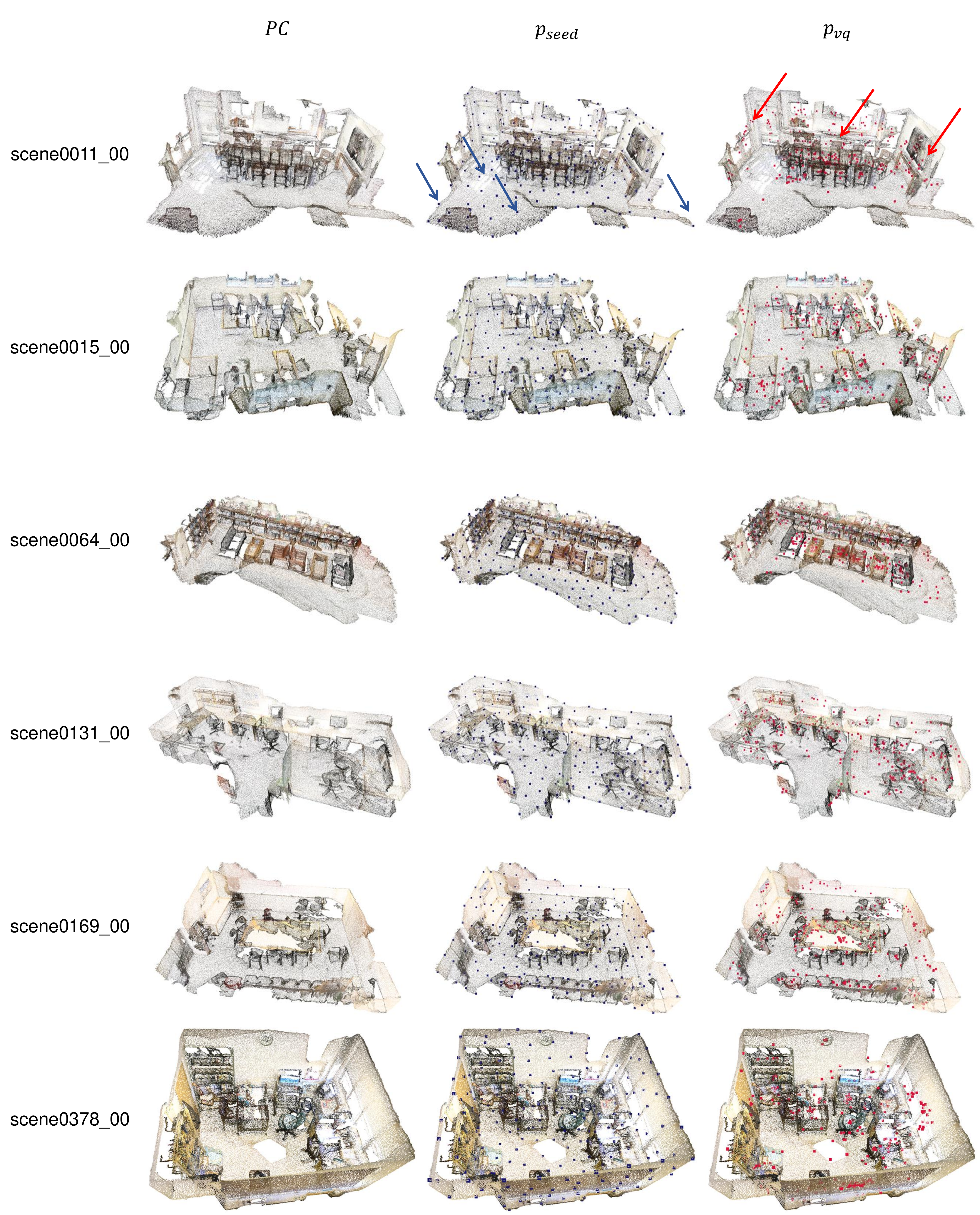}
	\caption{
	\textbf{Visualization of vote queries.} 
	We visualize absolute position of different object queries, $p_{seed}$ used in 3DETR (marked in blue) and $p_{vq}$ used in our proposed Vote2Cap-DETR (marked in red) with the input point cloud $PC$.
	Most of the vote queries focus on objects in a 3D scene (as red arrows pointed out), while $p_{seed}$ is mostly distributed in background areas (as blue arrows pointed out).
	}
	\label{fig:visualization-vote-query}
\end{figure*}

\textbf{Visualization results of vote queries.}
We visualize the vote queries' position $p_{vq}$ in our Vote2Cap-DETR and seed queries' position $p_{seed}$ of 3DETR in Figure \ref{fig:visualization-vote-query}.
Most of the vote queries focus on objects in a 3D scene, while $p_{seed}$ is mostly distributed in background areas.

\begin{figure*}[htbp]
	\centering
	\includegraphics[width=\linewidth]{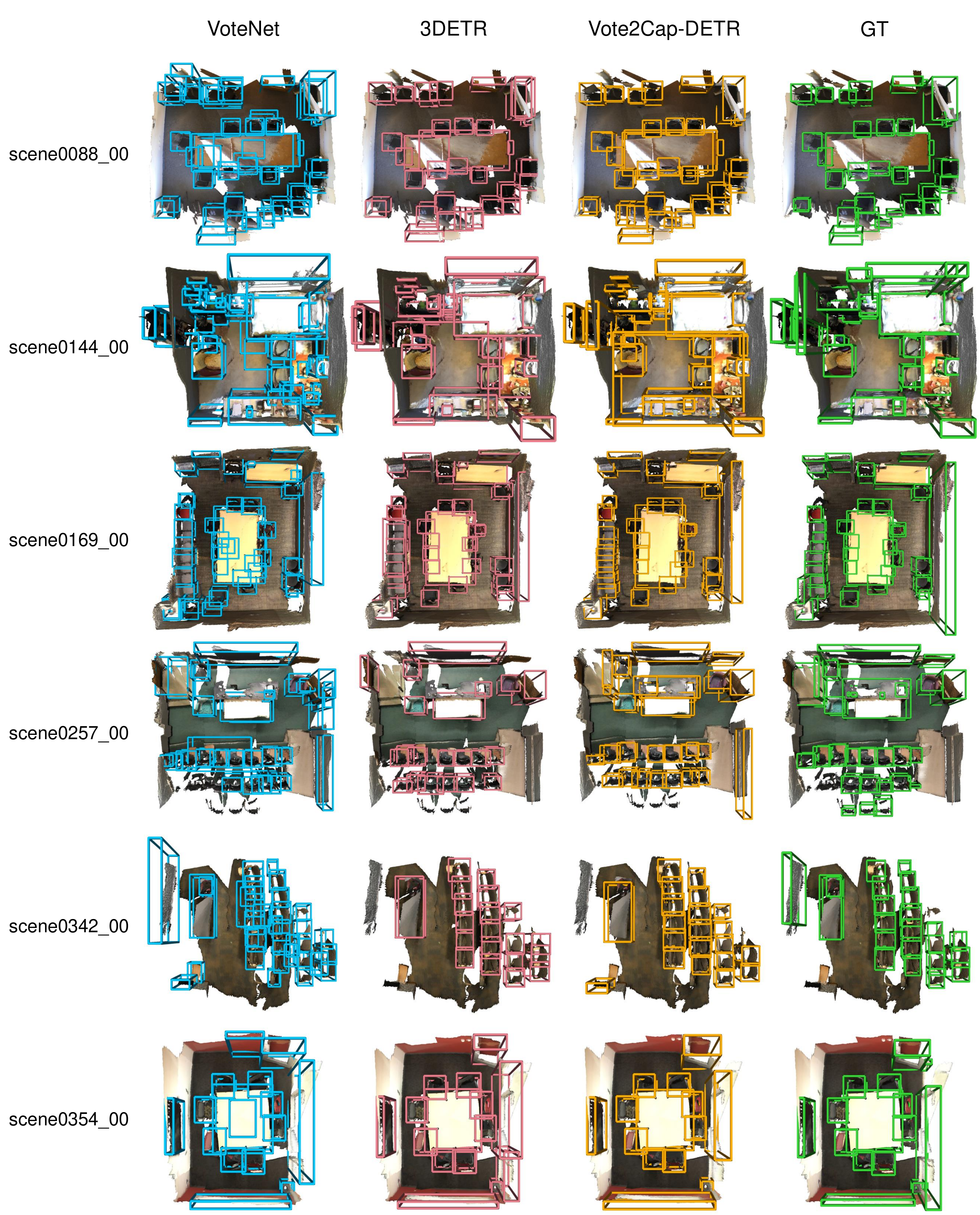}
	\caption{
	\textbf{Visualization of detection performance.} 
	We visualize detection results of VoteNet\cite{qi2019votenet}, 3DETR\cite{misra2021-3detr}, and our proposed Vote2Cap-DETR.
	Our proposed Vote2Cap-DETR is able to generate accurate localization results.
	}
	\label{fig:visualization-detection}
\end{figure*}

\textbf{Visualization of detection results.}
We visualize several detection results in Figure \ref{fig:visualization-detection}.
Our proposed Vote2Cap-DETR is able to generate accurate box predictions for a 3D scene.

\end{document}